\documentclass{ieeeaccess}
\usepackage{cite}
\usepackage{amsmath,amssymb,amsfonts}
\usepackage{algorithmic}
\usepackage{graphicx}
\usepackage{textcomp}
\usepackage{booktabs}
\newcommand{\cmark}{\ding{51}} 
\usepackage{makecell}
\usepackage{multirow}
\usepackage{pifont}
\usepackage{stfloats}

\def\BibTeX{{\rm B\kern-.05em{\sc i\kern-.025em b}\kern-.08em
    T\kern-.1667em\lower.7ex\hbox{E}\kern-.125emX}}

\begin{document}
\history{ }
\doi{ }

\title{Curriculum-Based Noise Adaptation for Phoneme-to-Text Reconstruction in Visual Speech Recognition}

\author{\uppercase{Matthew Kit Khinn Teng}, HAIBO ZHANG, and TAKESHI SAITOH}
\address{Graduate School of Computer Science and Systems Engineering, Kyushu Institute of Technology, Fukuoka 820-8502, Japan}

\tfootnote{This work was supported by Japan Society for the Promotion of Science (JSPS) KAKENHI Grant Number JP23H03787.}

\markboth
{MKK. Teng \headeretal: Curriculum-Based Noise Adaptation for Phoneme-to-Text Reconstruction in Visual Speech Recognition}
{MKK. Teng \headeretal: Curriculum-Based Noise Adaptation for Phoneme-to-Text Reconstruction in Visual Speech Recognition}

\corresp{Corresponding author: Matthew Kit Khinn Teng (e-mail: teng.khinn-matthew178@mail.kyutech.jp).}

\begin{abstract}
Phoneme-centric visual speech recognition reconstructs sentences from intermediate phoneme predictions, making overall recognition performance highly dependent on the robustness of the phoneme-to-text reconstruction model. Existing reconstruction approaches are commonly trained on clean phoneme sequences or synthetically corrupted inputs, leading to a mismatch between training conditions and the realistic phoneme prediction errors encountered during inference. To address this limitation, this paper proposes progressive error curriculum training (PECT). This curriculum learning framework progressively adapts a No Language Left Behind (NLLB)-based phoneme-to-text reconstruction model using synthetic phoneme perturbations, multi-domain pseudo-labels, and target-domain pseudo-labels generated by a visual speech recognizer. By gradually exposing the reconstruction model to increasingly realistic phoneme prediction errors, the proposed framework improves robustness while preserving sentence-reconstruction accuracy. Experiments on the LRS2 and LRS3 benchmarks demonstrate that PECT consistently improves reconstruction performance across multiple phoneme-based visual speech recognition frontends, including visual automatic speech recognition (V-ASR), point visual automatic speech recognition (PV-ASR), and head-pose-aware visual speech recognition (HP-VSR) variants. In particular, PECT reduces the word error rate (WER) of HP-VSR-FiLMFuse (L4) from 23.3\% to 22.2\% on LRS2 and reduces the WER of HP-VSR-ResFiLM from 30.3\% to 29.7\% on LRS3. Comprehensive ablation studies and qualitative analyses further demonstrate the effectiveness of progressively adapting the reconstruction model to realistic phoneme prediction errors. These results show that PECT provides an effective and generalizable curriculum learning strategy for phoneme-to-text reconstruction in phoneme-centric visual speech recognition.

\end{abstract}


\titlepgskip=-15pt

\maketitle

\section{Introduction} \label{sec:introduction}
\subsection{Background}

Visual speech Recognition (VSR), commonly referred to as lip-reading, aims to recognize spoken content using visual information alone. By relying exclusively on facial and lip movement cues, VSR provides a valuable alternative to conventional audio-based speech recognition in scenarios where acoustic signals are unavailable or severely degraded. Recent advances in deep learning, large-scale audio-visual datasets, and Transformer architectures have substantially improved VSR performance, enabling end-to-end recognition in challenging real-world environments.

Among recent developments \cite{teng_2026pvsr,teng_2026hpvsr,thomas_2025vallr}, phoneme-centric VSR frameworks have emerged as an effective alternative to direct word-level decoding. These approaches decompose recognition into two stages: a visual frontend that predicts phoneme sequences and a reconstruction module that converts phonemes into natural language text. This intermediate phoneme representation simplifies the visual-to-text mapping problem while enabling the integration of powerful pretrained language models. Recent studies have shown that multilingual encoder--decoder models such as No Language Left Behind (NLLB) \cite{costa_2022nllb} can effectively reconstruct sentences from phoneme sequences by treating phoneme-to-text conversion as a translation task, thereby improving sentence-level recognition performance.

Despite these advances, the phoneme-to-text reconstruction stage remains highly dependent on the quality of the phoneme predictions generated by the VSR. While reconstruction models are often trained on clean phoneme sequences or simple synthetic perturbations, practical VSR systems produce complex, context-dependent phoneme prediction errors. This discrepancy creates a gap between training and deployment conditions, limiting the effectiveness of existing reconstruction approaches. These observations motivate the need for training strategies that improve robustness to realistic phoneme prediction errors while preserving accurate phoneme-to-text translation capabilities.

\subsection{Problem and Motivation}
\label{motivation}

Although phoneme-centric architectures have demonstrated promising performance in VSR, several challenges remain unresolved in the sentence reconstruction stage. Recent studies have shown that phoneme sequences predicted by VSRs can be converted into natural-language text using large language models (LLMs) and neural machine translation architectures. For example, Thomas \textit{et al.}~\cite{thomas_2025vallr} employ a decoder-only language model to reconstruct sentences from predicted phoneme sequences, while our previous work \cite{teng_2026pvsr,teng_2026hpvsr} reformulates phoneme-to-text reconstruction as a translation task using the NLLB encoder--decoder model. Despite these advances, reconstruction performance remains highly dependent on the quality and error characteristics of the intermediate phoneme sequences generated by the VSR. To address these limitations, we identify the following key problems:

\begin{enumerate}

\item \textbf{Training--Inference Mismatch in Phoneme-to-Text Reconstruction.}

Existing phoneme-to-text reconstruction frameworks are primarily trained using clean phoneme sequences derived from ground-truth transcriptions. Under such settings, the reconstruction model learns ideal phoneme-to-word mappings and sentence-level linguistic relationships. During inference, however, the model receives phoneme predictions generated by a VSR system rather than ground-truth phonemes.

These predicted phoneme sequences inevitably contain recognition errors caused by visual ambiguities, coarticulation effects, motion blur, pose variations, and other challenges inherent to lip-reading. Consequently, a distribution mismatch arises between the clean phoneme sequences observed during training and the noisy phoneme predictions encountered during inference. This distinction can significantly degrade reconstruction performance, motivating the need for training strategies that better align reconstruction learning with realistic inference conditions.

\item \textbf{Limitations of Existing Noise Adaptation Strategies.}

Existing phoneme-to-text reconstruction approaches typically train reconstruction models using either clean phoneme-text pairs or synthetically corrupted phoneme sequences. Although synthetic error injection improves robustness to phoneme perturbations, the injected errors are usually generated through random insertion, deletion, and substitution operations that do not fully reflect the complex and context-dependent phoneme prediction errors produced by practical VSR systems. 

As a result, reconstruction models may become adapted to artificial noise distributions that differ substantially from the phoneme prediction errors encountered during deployment. This limitation suggests that robustness cannot be achieved through synthetic perturbations alone. Instead, effective adaptation should expose the reconstruction model to increasingly realistic error distributions, thereby motivating the integration of both synthetic phoneme perturbations and K-fold pseudo-labels generated by a VSR system within a unified curriculum learning framework.

\end{enumerate}

Motivated by these observations, we investigate a curriculum-based training strategy that gradually aligns phoneme-to-text reconstruction with the error distributions encountered during practical VSR inference.

\subsection{Our Contributions}
\label{contribution}

To address the limitations of existing phoneme-to-text reconstruction approaches, we propose a curriculum-based noise adaptation framework that progressively bridges the gap between clean phoneme supervision and realistic VSR prediction errors. The main contributions of this work are summarized as follows:

\begin{itemize}

\item \textbf{Progressive error curriculum training (PECT).}
We propose PECT, a curriculum learning framework that progressively trains the NLLB reconstruction model using synthetic phoneme perturbations, multi-domain pseudo-labels, and target-domain pseudo-labels. By gradually exposing the model to increasingly realistic phoneme prediction errors, PECT reduces the mismatch between training and inference, resulting in more robust phoneme-to-text reconstruction.

\item \textbf{Target-domain pseudo-label adaptation with K-fold supervision.}
We introduce a K-fold pseudo-label generation strategy that produces out-of-fold phoneme predictions for curriculum training without data leakage. By exposing the reconstruction model to realistic target-domain phoneme error distributions, the proposed strategy provides more representative supervision than synthetic perturbations alone and further improves robustness during inference.

\item \textbf{Comprehensive experimental evaluation.}
We conduct extensive experiments on the LRS2 and LRS3 benchmarks, including curriculum ablation studies, comparisons with state-of-the-art phoneme-centric VSR methods, and cross-model evaluations using multiple Stage~1 VSR frontends. The results demonstrate that PECT improves phoneme-to-text reconstruction and generalizes across different phoneme-based VSR architectures without requiring architecture-specific retraining.

\end{itemize}

\section{Related Works}
\label{sec:related_works}

\subsection{Phoneme- and Viseme-Centric Visual Speech Recognition}

VSR has evolved from early spatio-temporal convolutional and recurrent architectures to Transformer-based models capable of capturing long-range temporal dependencies. Large-scale datasets such as LRS2 \cite{son_2017lrs2} and LRS3 \cite{afouras_2018lrs3} have facilitated the development of increasingly powerful visual encoders, while recent self-supervised approaches such as AV-HuBERT \cite{shi_2022avhubert} have further improved representation learning from unlabeled audio-visual data \cite{shi_2022interspeech}. More recently, phoneme-centric VSR frameworks have gained attention by introducing intermediate linguistic representations between visual features and sentence-level outputs. Rather than directly predicting words, these methods first estimate phoneme sequences and subsequently reconstruct sentences using pronunciation lexicons or neural language models, thereby decoupling visual recognition from linguistic decoding and improving linguistic generalization \cite{shillingford_2019interspeech,prajwal_2022subword,thomas_2025vallr}. This formulation also enables the integration of powerful language models for downstream phoneme-to-text reconstruction.

A fundamental challenge in visual speech recognition is the many-to-one relationship between phonemes and visemes. Multiple phonemes often share nearly identical lip movements and therefore map to the same viseme, making them visually indistinguishable even for an ideal recognition model \cite{bear_2017SC}. Consequently, phoneme prediction inevitably contains ambiguities arising from the visual speech signal itself. These inherent uncertainties motivate phoneme-centric architectures that leverage sentence-level linguistic context to recover the intended word sequence from imperfect phoneme predictions.

Early efforts in this direction investigated two-stage decoding frameworks based on viseme or phoneme representations. Peymanfard \textit{et al.}~\cite{peymanfard_2022ieee} proposed a two-stage architecture that first converts visual speech into viseme sequences and subsequently decodes them into characters using a separate language model. By leveraging external text corpora for viseme-to-character conversion, their method achieved a 4\% absolute reduction in word error rate (WER) over conventional sequence-to-sequence baselines on LRS2. Similarly, El-Bialy \textit{et al.}~\cite{elBialy_CAAI_2023} investigated phoneme-based lip reading using a spatio-temporal visual encoder, a Transformer-based phoneme recognizer, and a recurrent neural network language model for sentence generation. Their results demonstrated that phoneme-level supervision outperformed both character- and viseme-based alternatives, yielding substantial improvements on the LRS2 benchmark.

These findings motivated more recent phoneme-centric VSR systems that replace pronunciation lexicons or conventional language models with large language models and neural machine translation architectures for phoneme-to-text reconstruction, including VALLR \cite{thomas_2025vallr} and subsequent NLLB-based frameworks \cite{teng_2026pvsr,teng_2026hpvsr}. Unlike earlier approaches that primarily focused on improving phoneme recognition, these methods emphasize robust sentence reconstruction from predicted phoneme sequences. In this work, we build upon this paradigm and focus on improving the robustness of the phoneme-to-text reconstruction stage through curriculum-based training.

\subsection{LLMs for VSR}

Recent advances in sequence-to-sequence learning and LLMs have significantly influenced the development of speech recognition and visual speech recognition systems. Encoder--decoder architectures, exemplified by Listen, Attend and Spell \cite{chan_2016icassp} and later speech frameworks such as ESPnet \cite{watanabe_2018espnet}, demonstrated the effectiveness of jointly modeling acoustic representations and linguistic context within a unified sequence-to-sequence framework. Building upon these advances, recent LLM-based approaches have extended pretrained language models beyond text processing to speech understanding and generation, as demonstrated by SpeechGPT \cite{zhang_2023speechgpt}. In the VSR domain, Prajwal \textit{et al.} \cite{prajwal_2024interspeech} showed that pretrained Audio Speech Recognition (ASR) models such as Whisper \cite{radford_2023whisper} can be adapted for lip reading through cross-modal alignment, while Cappellazzo \textit{et al.} \cite{cappellazzo_2025llama} proposed Llama-AVSR, which combines modality-specific encoders with an Llama-based decoder for multimodal speech recognition. More closely related to our work, Thomas \textit{et al.} \cite{thomas_2025vallr} introduced a phoneme-centric Visual Automatic Speech Recognition (V-ASR) framework that predicts phoneme sequences from visual inputs and reconstructs text using a language model decoder.

Beyond direct speech recognition, recent studies have also explored adapting pretrained sequence-to-sequence language models for downstream speech processing tasks, including automatic speech recognition error correction and text restoration from noisy intermediate representations. These studies demonstrate that large pretrained language models can effectively recover linguistic information from imperfect recognition outputs by exploiting contextual knowledge. In parallel, advances in multilingual machine translation have produced large-scale encoder-decoder models with strong cross-lingual transfer capabilities. Among these, NLLB \cite{costa_2022nllb} introduced a massively multilingual sequence-to-sequence model capable of supporting hundreds of languages, including many low-resource and underrepresented languages. Owing to its strong cross-lingual transfer capabilities, NLLB has been adapted beyond conventional machine translation tasks. Teng \textit{et al.} \cite{teng_2026pvsr} reformulated phoneme-to-text reconstruction as a translation problem by treating phoneme sequences as a pseudo-low-resource language and translating them into natural language sentences using a fine-tuned NLLB model. This approach leverages the multilingual knowledge embedded in NLLB to recover linguistic information from phoneme predictions and shows improved reconstruction performance compared with conventional decoding approaches.

Building on these studies, our work examines how to improve robustness to phoneme prediction errors through curriculum-based adaptation of the NLLB reconstruction model using synthetic phoneme errors and K-fold pseudo-labels generated from a VSR system.

\begin{figure*}[tb]
\centerline{\includegraphics[width=\textwidth]{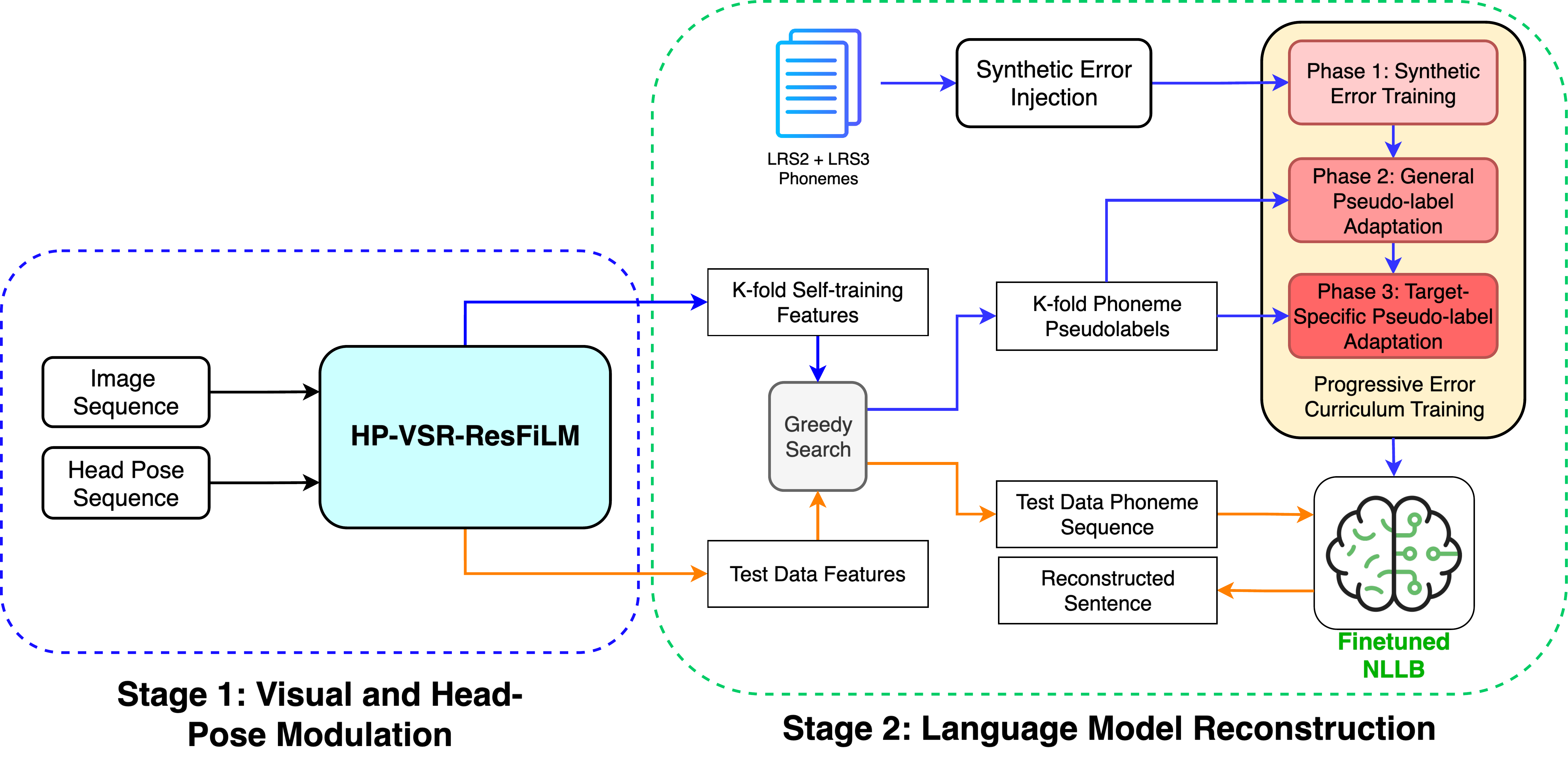}}
\caption{Overview of the proposed PECT framework. The training pipeline has three sequential curriculum phases. Phase~1 uses synthetic phoneme perturbations with a synthetic error injector to improve robustness to recognition errors. Phase~2 carries out multi-domain pseudo-label adaptation with K-fold pseudo-labels generated from the combined LRS2 and LRS3 datasets, allowing the reconstruction model to adapt to realistic phoneme prediction errors. Phase~3 performs target-domain pseudo-label adaptation with K-fold pseudo-labels generated from the target dataset (LRS2 or LRS3), further refining the model for the target domain. The resulting fine-tuned NLLB model is then used at inference time to reconstruct natural-language text from phoneme sequences predicted by the Stage~1 VSR frontend. Solid arrows denote the training pipeline, while dashed arrows indicate the inference pathway.}
\label{fig:overall_architecture}
\end{figure*}

\subsection{Curriculum and Robust Phoneme-to-Text Reconstruction}
\subsubsection{Synthetic Error Injection}
Noise-aware training strategies have been explored to improve the robustness of phoneme-based recognition and translation systems. Lee~\textit{et~al.}~\cite{lee_2023asru} proposed a two-pass cross-lingual transfer learning framework for phoneme recognition and phoneme-to-grapheme translation, combining triphone pseudo-noise labeling with a transformer-based global phoneme noise generator to simulate realistic phoneme recognition errors. Although their framework also incorporates pseudo-label refinement, its primary objective is to improve low-resource acoustic speech recognition through learned phoneme noise generation. In contrast, our work targets the visually induced phoneme prediction errors encountered in VSR, which differ substantially from those in acoustic speech recognition.

More broadly, robustness to recognition errors has attracted increasing attention in downstream language processing tasks. Existing studies have explored synthetic error generation, data augmentation, and noisy intermediate representation modeling to improve resilience against ASR-induced errors \cite{wang_2020nlpcai,cui_2021asr,nakagome2022interaug,binici2025medsage}. These approaches expose language models to simulated insertion, deletion, and substitution errors, as well as noisy intermediate predictions, during training, enabling more robust performance under realistic recognition conditions. More fundamentally, exposure bias arising from the mismatch between clean training inputs and noisy inference conditions has been identified as a major challenge for sequence generation models~\cite{bengio_2015scheduled}. Likewise, deep neural networks are highly sensitive to noisy supervision, motivating progressive adaptation to increasing levels of label uncertainty~\cite{song_2023ieee}.

Despite these advances, existing methods still rely mainly on synthetic perturbations from acoustic speech recognition and fail to capture the distinctive phoneme prediction errors seen in VSR. In addition, synthetic error injection is typically used as a static augmentation strategy rather than as part of a progressive learning framework. To address these limitations, we combine a lightweight VSR-specific phoneme error injection strategy with K-fold pseudo-labels in a three-phase curriculum that progressively adapts a multilingual NLLB model to realistic phoneme prediction errors.

\subsubsection{Curriculum Learning}

Curriculum learning improves optimization and model generalization by presenting training samples in a meaningful 
order from easier to more difficult examples~\cite{bengio_2009curriculum}. Theoretical analyses have further demonstrated that curriculum learning is most beneficial when the training distribution gradually approaches the target task distribution, providing a principled motivation for structured noise exposure~\cite{hacohen2019power}. Self-paced learning extended this paradigm by allowing the model to automatically select reliable training samples during optimization, improving robustness to difficult or noisy examples~\cite{kumar2010selfpaced}. Self-paced curriculum learning further combined manually designed curricula with adaptive sample selection, enabling progressively more challenging training while accounting for model confidence~\cite{jiang2015selfpaced}. Automated curriculum learning subsequently removed the need for manual curriculum design by dynamically adjusting task difficulty during optimization, demonstrating improved convergence in deep neural networks~\cite{graves2017automated}. A comprehensive survey of curriculum learning methods further established its broad applicability across computer vision, speech recognition, and natural language processing, highlighting its consistent benefits for optimization stability and model robustness~\cite{soviany2022curriculum}.

Several lines of research have explored techniques closely related to our proposed curriculum-based training framework, though these ideas have largely been investigated independently. In the context of noise-robust Natural Language Processing (NLP), Hu~\textit{et~al.}~\cite{shi2024robustger} proposed RobustGER, extending generative error correction to noisy ASR conditions by fine-tuning LLMs to map noisy speech hypotheses to correct transcriptions, demonstrating that LLMs can be effectively adapted to handle recognition errors in downstream text generation. Separately, curriculum learning has been applied to noisy text generation tasks, where Hari~\textit{et~al.}~\cite{hari2024curriculum} demonstrated that ordering training samples from clean to noisy data using an annealing schedule yields measurable improvements in cross-lingual data-to-text generation, highlighting the benefit of structured noise exposure during training.

Although these studies motivate individual components of our approach, their integration into a unified framework for robust phoneme-to-text reconstruction in phoneme-centric VSR has not been investigated previously. Moreover, unlike conventional noise-robust NLP methods that operate on word-level recognition hypotheses, our work addresses phoneme-level prediction errors produced by VSR systems, where visual ambiguities create distinct error patterns. To bridge this gap, we propose a curriculum learning framework that progressively adapts the reconstruction model to increasingly realistic phoneme prediction errors.

\subsubsection{Pseudo-Label Learning}

Pseudo-label learning has become a widely adopted semi-supervised learning strategy for leveraging unlabeled data by treating model predictions as supervisory signals. Representative approaches such as Noisy Student~\cite{xie_2020noisystudent}, FixMatch~\cite{sohn_2020fixmatch}, Cross Pseudo Supervision~\cite{chen_2021cps}, and SoftTeacher~\cite{xu_2021softteacher} have demonstrated that high-quality pseudo-labels can substantially improve model generalization across image classification, semantic segmentation, and object detection. These methods iteratively refine model predictions or employ teacher--student learning to provide increasingly reliable supervision from unlabeled data.

Pseudo-label learning has also been widely explored in automatic speech recognition, where unlabeled speech is incorporated through iterative self-training and confidence-based pseudo-label refinement. Jin~\textit{et~al.}~\cite{jin_2022filterevolve} proposed a progressive pseudo-label refinement strategy that iteratively filters low-confidence predictions and updates the training set with increasingly reliable pseudo-labels, improving semi-supervised speech recognition performance. These studies demonstrate the effectiveness of progressive pseudo-label learning for acoustic speech recognition.

To the best of our knowledge, pseudo-label learning has not been investigated for phoneme-to-text reconstruction in phoneme-centric VSR. Unlike acoustic speech recognition, VSR produces phoneme prediction errors due to visual ambiguities, such as confusable viseme pairs and head-pose variations, resulting in error characteristics that differ substantially from those of ASR-induced noise. Our work addresses this gap by incorporating K-fold VSR-generated pseudo-labels into a curriculum learning framework for robust phoneme-to-text reconstruction.

\section{Methodology}
\label{sec:methodology}

Figure~\ref{fig:overall_architecture} illustrates the overall framework of the proposed curriculum-based phoneme-to-text reconstruction system. The framework consists of two stages. In Stage~1, a VSR model predicts phoneme sequences from visual speech inputs. These phoneme predictions are utilized both for final evaluation and for generating K-fold pseudo-labels that simulate realistic recognition errors. In Stage~2, an NLLB-based phoneme-to-text reconstruction model is trained using a progressive curriculum learning strategy. Training begins with synthetically perturbed phoneme sequences, followed by general and target-specific adaptation with pseudo-labels generated by the Stage~1 VSR frontend. This progressive curriculum exposes the reconstruction model to increasingly realistic phoneme prediction errors, reducing the mismatch between training and inference and improving sentence reconstruction under realistic VSR conditions.

\subsection{Stage 1: Visual and Head-Pose Modulation}

The proposed framework utilizes phoneme sequences generated by the HP-VSR-ResFiLM model introduced in our previous work \cite{teng_2026hpvsr}. During inference, phoneme predictions produced by the Stage~1 VSR frontend serve as the input to the Stage~2 phoneme-to-text reconstruction module.

To expose the reconstruction model to realistic recognition errors during training, we employ a five-fold pseudo-label generation strategy. The training data are partitioned into five mutually exclusive folds, and out-of-fold phoneme predictions are generated for all training samples. These pseudo-labels contain naturally occurring phoneme prediction errors produced by the VSR system and therefore provide a closer approximation to the phoneme sequences encountered during inference than synthetically generated noise.

By utilizing out-of-fold predictions, the pseudo-label generation process prevents data leakage between training samples and their corresponding phoneme predictions. The resulting pseudo-labels are subsequently used during the final stage of the proposed curriculum learning framework to adapt the reconstruction model to realistic VSR errors.


\subsection{Stage 2: Language Model Reconstruction}
Stage~2 aims to reconstruct natural language sentences from phoneme sequences generated by the VSR model. To improve robustness against phoneme prediction errors, we introduce a curriculum-based training framework consisting of a synthetic error injector, PECT module, and an NLLB-based phoneme-to-text reconstruction model.

\subsubsection{Synthetic Error Injector} \label{injector}
To improve the robustness of the phoneme-to-text reconstruction model to recognition errors, we introduce a synthetic error injector that simulates common phoneme prediction mistakes produced by VSR systems. Given a clean phoneme sequence, the injector randomly applies insertion, deletion, and substitution operations to generate a corrupted phoneme sequence. These operations are designed to mimic the error patterns typically observed in phoneme predictions, where phonemes may be omitted, incorrectly recognized, or spuriously inserted.

For each phoneme sequence, a noise ratio of 20\% is applied. Specifically, approximately 20\% of the phoneme positions within a sequence are selected at random and modified using one of the three error operations. Insertion introduces an additional phoneme token at the selected position, deletion removes the corresponding phoneme token, and substitution replaces the original phoneme with another phoneme sampled from the ARPAbet vocabulary. The error type is selected randomly with equal probability. The generated noisy phoneme sequences are subsequently utilized during Phase~1 of the proposed PECT.

\subsubsection{PECT Module}

Curriculum learning aims to improve model optimization by presenting training samples in a meaningful order, typically progressing from easier to more difficult examples \cite{bengio_2009curriculum}. Motivated by this principle, we propose a PECT strategy that gradually exposes the phoneme-to-text reconstruction model to increasing levels of phoneme prediction noise. Rather than training directly on noisy phoneme sequences, the model is first optimized on clean phoneme-text pairs and subsequently adapted using synthetic and realistic phoneme errors. This progressive learning process enables the model to establish strong phoneme-to-text mappings before learning to recover from recognition errors commonly produced by VSR systems.

\textbf{Phase 1: Synthetic error training.}
In the first phase, clean phoneme sequences are processed using the synthetic error injector described in Section~\ref{injector}. The resulting noisy phoneme sequences are paired with their original sentence transcriptions and used for training. By introducing insertion, deletion, and substitution errors, this stage exposes the reconstruction model to controlled phoneme perturbations while maintaining correct sentence-level supervision. As a result, the model learns to recover linguistic content from imperfect phoneme inputs and develops robustness to recognition errors.

\textbf{Phase 2: Multi-domain pseudo-label adaptation.}
In the second phase, the model is further trained using pseudo-labels generated by the Stage~1 VSR system on the combined LRS2 and LRS3 datasets. Compared with synthetic perturbations, these pseudo-labels reflect realistic recognition errors from actual visual speech recognition, including context-dependent substitutions, insertions, and deletions. This stage enables the reconstruction model to adapt to the statistical characteristics of real VSR outputs while benefiting from the linguistic and visual diversity of both datasets.

\textbf{Phase 3: Target-domain pseudo-label adaptation.}
In the final phase, the model is refined using K-fold pseudo-labels generated from the target dataset. This stage further aligns the reconstruction model with the target-domain vocabulary, speaking style, and phoneme error distribution encountered during inference. By adapting to target-specific recognition patterns, the model reduces the remaining mismatch between training and testing conditions and enhances reconstruction accuracy.

Unlike clean phoneme supervision, which encourages reliance on nearly error-free inputs, the proposed curriculum progressively exposes the reconstruction model to increasingly realistic phoneme prediction errors. Beginning with synthetic perturbations and subsequently transitioning to source-domain and target-domain pseudo-labels enables the model to learn robust phoneme-to-text mappings under conditions that closely resemble those encountered at inference time. Consequently, the reconstruction model becomes more effective at recovering sentence-level semantics from noisy phoneme sequences, thereby improving robustness and reconstruction performance.

\subsubsection{NLLB Phoneme-to-Text Reconstruction Model}

We adopt the NLLB Transformer encoder--decoder architecture \cite{costa_2022nllb} as the backbone for phoneme-to-text reconstruction. Following our previous work \cite{teng_2026pvsr}, phoneme sequences are treated as a pseudo-low-resource language and translated into English sentences using the multilingual translation capabilities of NLLB. The encoder transforms the input phoneme sequence into contextualized representations through stacked self-attention and feed-forward layers. At the same time, the decoder autoregressively generates the target sentence by attending to both the encoder representations and previously generated tokens. A linear projection layer followed by a softmax operation produces probability distributions over the target vocabulary at each decoding step, and the final output sentence is obtained by selecting the most likely token at each step during generation. Owing to its strong cross-lingual transfer capability, NLLB provides an effective foundation for recovering natural language text from imperfect phoneme sequences, particularly when the input contains recognition errors, omitted phonemes, or other perturbations introduced by the VSR frontend.

\section{Experiments}
\label{sec:experiments}

\subsection{Dataset}
\label{datasets}
Experiments were conducted on the LRS2 and LRS3 benchmarks, which are widely used for large-vocabulary VSR. LRS2 \cite{son_2017lrs2} contains approximately 224.5 hours of sentence-level audio-visual speech collected from BBC television broadcasts, comprising over 144,000 utterances. LRS3 \cite{afouras_2018lrs3} is currently the largest publicly available English audio-visual speech dataset, consisting of approximately 439 hours of speech extracted from TED and TEDx talks and containing more than 151,000 utterances. Both datasets exhibit substantial variability in speaker identity, pose, illumination, and speaking style, making them suitable benchmarks for evaluating phoneme-based VSR and phoneme-to-text reconstruction systems.

\subsection{Preprocess}
\label{preprocess}

\textbf{Video and head pose preprocessing.} 
The VSR frontend and video preprocessing pipeline follow our previous work \cite{teng_2026hpvsr,teng_2026pvsr}. Specifically, face detection, mouth-region cropping, frame normalization, and head-pose estimation are performed using the same preprocessing procedures and parameter settings. Readers are referred to \cite{teng_2026hpvsr,teng_2026pvsr} for the detailed implementation information.

\textbf{Phoneme labels preprocessing.} 
Following our previous phoneme-based VSR framework \cite{teng_2026hpvsr,teng_2026pvsr}, all utterances from the LRS2 \cite{son_2017lrs2} and LRS3 \cite{afouras_2018lrs3} datasets are converted into phoneme-text pairs for training and evaluation. Ground-truth transcriptions are first normalized using the NVIDIA NeMo text normalization toolkit \cite{zhang_2024nemo} and subsequently converted into context-aware ARPAbet phoneme sequences using the SoundChoice grapheme-to-phoneme converter \cite{ploujnikov_2022soundchoice}. The resulting phoneme vocabulary consists of 39 phoneme classes, consistent with prior phoneme-based VSR studies \cite{teng_2026hpvsr,teng_2026pvsr}. 

\textbf{K-fold pseudolabels generation.} 
To generate realistic phoneme prediction errors for curriculum training, we adopt a five-fold pseudo-labeling strategy. The combined pretraining and training partitions of the LRS2 and LRS3 datasets are divided into five mutually exclusive folds, each containing approximately equal numbers of video clips. Specifically, each fold contains approximately 28,740 clips for LRS2 and 32,980 clips for LRS3. Partitioning is performed at the video level to prevent overlap between training and pseudo-label generation data, thereby minimizing information leakage across folds.

\subsection{Experimental Setup}
\label{exp_setup}

\textbf{Stage 1 VSR configuration.}
The Stage~1 phoneme test data predictions are generated using the HP-VSR-ResFiLM pretrained model proposed in our previous work \cite{teng_2026hpvsr}. For K-fold pseudo-label generation, the model is initialized using the publicly available English pretrained weights from Auto-AVSR \cite{ma_2023avsr}. We adopt the same architecture and training configuration as \cite{teng_2026hpvsr} to ensure consistency between pseudo-label generation and final evaluation. 

\textbf{K-fold pseudo-label generation.}
The combined pretraining and training partitions of the LRS2 and LRS3 datasets are divided into five mutually exclusive folds containing approximately equal numbers of video clips. For each fold, an HP-VSR-ResFiLM model is initialized using the English pretrained weights from Auto-AVSR \cite{ma_2023avsr} and trained on the remaining four folds. Training is performed for a maximum of 80 epochs with an initial learning rate of $1\times10^{-4}$ and early stopping after five consecutive epochs without improvement on the validation set. After training, the model is used to generate phoneme predictions for the held-out fold. This process is repeated for all five folds, and the resulting out-of-fold phoneme predictions are aggregated to form the pseudo-labeled dataset used during Phase~3 of curriculum training. To prevent information leakage, no validation or test samples from either LRS2 or LRS3 are used during pseudo-label generation.

\textbf{NLLB training configuration.}
The NLLB reconstruction model is trained using the AdamW optimizer with a batch size of 16 and an initial learning rate of $5\times10^{-5}$. Curriculum learning is performed sequentially across three phases, with each phase trained for 10 epochs. The best-performing checkpoint from each phase is used to initialize the subsequent phase. To facilitate stable adaptation to increasingly challenging phoneme inputs, the learning rate is reduced by a factor of 10 at each curriculum transition. Consequently, Phase~1, Phase~2, and Phase~3 are trained using learning rates of $5\times10^{-5}$, $5\times10^{-6}$, and $5\times10^{-7}$, respectively. 

\begin{table*}[t]
\centering
\caption{Comparison of visual speech recognition methods on the LRS2 and LRS3 datasets. The table shows whether each method uses phoneme-based recognition, an LLM, curriculum learning for phoneme-to-text reconstruction, external training data, and the amount of target-domain training data used during supervised adaptation. Training Data (h) denotes the duration of labeled LRS2/LRS3 data used for fine-tuning and evaluation. Extra Data indicates whether additional datasets beyond the target benchmarks are used during pretraining or training. Our HP-VSR variants are initialized from Auto-AVSR pretrained weights trained on 3,448 hours of large-scale audio-visual data and are fine-tuned only on the standard LRS2 or LRS3 training sets. PECT-NLLB denotes the proposed Phoneme Error Curriculum Training strategy applied to the NLLB phoneme-to-text reconstruction model.}
\label{tab:sota_curriculum}
\resizebox{\linewidth}{!}{
\begin{tabular}{lccccccccc}
\toprule
\multirow{2}{*}{\textbf{Method}} &
\multirow{2}{*}{\makecell{\textbf{Phoneme}\\\textbf{-based}}} &
\multirow{2}{*}{\textbf{LLM}} &
\multirow{2}{*}{\makecell{\textbf{Phoneme}\\\textbf{Curriculum}}} &
\multirow{2}{*}{\makecell{\textbf{Extra}\\\textbf{Data}}} &
\multicolumn{2}{c}{\textbf{LRS2}} &
\multicolumn{2}{c}{\textbf{LRS3}} \\
\cmidrule(lr){6-7} \cmidrule(lr){8-9} & & & & &
\makecell{\textbf{Training}\\\textbf{Data [h]}} &
\textbf{WER (\%)} &
\makecell{\textbf{Training}\\\textbf{Data [h]}} &
\textbf{WER (\%)} \\
\midrule

Auto-AVSR \cite{ma_2023avsr} & -- & -- & -- & \cmark & 818 & 27.9 & 818 & 33.0 \\
Auto-AVSR \cite{ma_2023avsr} & -- & -- & -- & \cmark & 3448 & 14.6 & 3448 & 19.1 \\
VALLR \cite{thomas_2025vallr} & \cmark & LLaMA & -- & -- & 28 & 20.8 & 30 & 18.7 \\
V-ASR \cite{teng_2026pvsr} & \cmark & NLLB (1.3B) & -- & -- & 223 & 35.2 & 438 & 41.2 \\
PV-ASR \cite{teng_2026pvsr} & \cmark & NLLB (1.3B) & -- & -- & 223 & 26.9 & 438 & 38.7 \\
Auto-AVSRFine-tunedd Baseline \cite{teng_2026hpvsr} & -- & -- & -- & -- & 223 & 33.0 & 438 & 30.7 \\ 
HP-VSR-Base \cite{teng_2026hpvsr} & \cmark & NLLB (1.3B) & -- & -- & 223 & 33.3 & 438 & 40.8 \\
HP-VSR-FiLMFuse (L4) \cite{teng_2026hpvsr} & \cmark & NLLB (1.3B) & -- & -- & 223 & 23.3 & 438 & 33.0 \\
HP-VSR-ResFiLM \cite{teng_2026hpvsr} & \cmark & NLLB (1.3B) & -- & -- & 223 & 24.7 & 438 & 30.3 \\
\midrule
HP-VSR-Base \cite{teng_2026hpvsr} & \cmark & PECT-NLLB (Ours) & \cmark & -- & 223 & 29.1 & 438 & 36.1 \\
HP-VSR-FiLMFuse (L4) \cite{teng_2026hpvsr} & \cmark & PECT-NLLB (Ours) & \cmark & -- & 223 & \textbf{22.2} & 438 & 31.3 \\
HP-VSR-ResFiLM \cite{teng_2026hpvsr} & \cmark & PECT-NLLB (Ours) & \cmark & -- & 223 & 23.6 & 438 & \textbf{29.7} \\
\bottomrule
\end{tabular}
}
\end{table*}

\subsection{Evaluation Metric}
\label{eval_metric}

To evaluate the effectiveness of the proposed phoneme-to-text reconstruction framework, we adopt word error rate (WER)~\cite{jelinek_1975wer} as the primary evaluation metric. WER measures the discrepancy between the reconstructed sentence and the reference transcription by accounting for substitution, deletion, and insertion errors. It is defined as

\begin{equation}
\text{WER} = \frac{S + D + I}{N},
\label{eq:wer}
\end{equation}

\noindent
where $S$, $D$, and $I$ denote the numbers of substitutions, deletions, and insertions, respectively, and $N$ is the total number of words in the reference transcription.

In addition, we report phoneme error rate (PER) to evaluate the quality of the phoneme sequences predicted by the Stage~1 visual speech recognizer. PER is computed using the same edit-distance formulation as WER, with words replaced by phonemes. Specifically, $S$, $D$, and $I$ represent phoneme substitutions, deletions, and insertions, while $N$ denotes the total number of reference phonemes. Whereas WER evaluates the final sentence reconstruction performance, PER measures the accuracy of the intermediate phoneme predictions. Reporting both metrics enables analysis of the relationship between phoneme recognition accuracy and downstream text reconstruction performance.

\section{Results}
\label{sec:results}
The proposed PECT consistently improves the performance of the phoneme-centric HP-VSR framework on both the LRS2 and LRS3 benchmarks, as shown in Table ~\ref{tab:sota_curriculum}. Compared with the original NLLB reconstruction model, PECT reduces the WER of HP-VSR-FiLMFuse (L4) from 23.3\% to 22.2\% on LRS2 and improves HP-VSR-ResFiLM from 30.3\% to 29.7\% on LRS3. These results demonstrate that progressively adapting the reconstruction model using synthetic phoneme perturbations and realistic pseudo-labels enables more robust phoneme-to-text reconstruction than conventional NLLB training.

The largest improvement is observed for HP-VSR-Base, where the WER decreases from 33.3\% to 29.1\% on LRS2 and from 40.8\% to 36.1\% on LRS3. This indicates that the proposed curriculum is particularly effective when the Stage~1 VSR frontend produces relatively noisy phoneme predictions. As the quality of the phoneme predictions improves, the reconstruction model has fewer opportunities to correct recognition errors, resulting in smaller but consistent gains for HP-VSR-FiLMFuse and HP-VSR-ResFiLM. These findings suggest that PECT complements stronger phoneme recognition frontend while remaining effective across all evaluated HP-VSR variants.

Compared with previous phoneme-centric approaches, the proposed HP-VSR variants substantially outperform V-ASR and PV-ASR on both benchmarks. Relative to PV-ASR, PECT-NLLB achieves a 4.7 percentage-point absolute WER reduction on LRS2 (26.9\% → 22.2\%) and a 9.0 percentage-point reduction on LRS3 (38.7\% → 29.7\%). These gains indicate that improving the phoneme-to-text reconstruction stage is complementary to advances in the VSR frontend.

Although VALLR reports lower WER than the proposed approach, it employs a large LLaMA-based language model together with additional external data, whereas the proposed framework is trained using only the standard LRS2/LRS3 target datasets during adaptation. Likewise, Auto-AVSR pretrained on 3,448 hours of audio-visual data achieves the best overall performance using direct word-level recognition, illustrating the benefit of large-scale pretraining. In contrast, the proposed framework focuses on phoneme-centric VSR, where sentence reconstruction is performed from predicted phoneme sequences. Despite this more challenging setting, the proposed PECT-NLLB model significantly narrows the performance gap while remaining competitive without requiring additional adaptation data.

\section{Ablation Study}
\label{sec:ablation_study}

\subsection{Curriculum Training}

\begin{table*}[t]
\centering
\caption{Effect of different training schedules on visual speech recognition performance. $C$, $S$, and $P$ denote clean phoneme supervision, 20\% synthetic-error phoneme supervision, and K-fold pseudo-label supervision, respectively. The subscript ``23'' indicates training on the combined LRS2+LRS3 dataset, while the subscript ``t'' denotes adaptation on the target dataset (LRS2 or LRS3). The notation ``$\rightarrow$'' indicates sequential curriculum training across stages. ``Single-stage'' denotes direct training on the target dataset without curriculum pretraining. ``$M_t$'' denotes uniform sampling from clean, synthetic-error, and pseudo-labeled phoneme sequences. V-ASR, PV-ASR, Base, ResFiLM, and FiLMFuse denote the Stage~1 visual speech recognition models used to generate phoneme predictions via greedy decoding, where Base, ResFiLM, and FiLMFuse correspond to the HP-VSR variants.}
\label{tab:curriculum}
\resizebox{\linewidth}{!}{
\begin{tabular}{l|cccccc|cccccc}
\toprule
\multirow{2}{*}{Training Schedule}
& \multicolumn{6}{c|}{LRS2, WER (\%)}
& \multicolumn{6}{c}{LRS3, WER (\%)} \\
\cmidrule(lr){2-7}
\cmidrule(lr){8-13}
& V-ASR & PV-ASR & Base & ResFiLM & \makecell{FiLMFuse\\(L3--4)} & \makecell{FiLMFuse\\(L4)}
& V-ASR & PV-ASR & Base & ResFiLM & \makecell{FiLMFuse\\(L3--4)} & \makecell{FiLMFuse\\(L4)}
\\
\midrule
$C_t$ (single-stage) & 34.67 & 26.92  & 35.75 & 25.11 & 26.01 & 24.23 & \textbf{37.80} & 35.34 & 41.29 & 30.15 & 32.08 & 32.56 \\
$S_t$ (single-stage) & 35.06 & 27.30 & 34.52 & 25.38 & 25.95 & 24.13 & 38.14 & 35.69 & 40.61 & 30.16 & 32.26 & 32.63 \\
$P_t$ (single-stage) & 34.61 & 25.86 & 31.86 & 25.32 & 25.08 & 23.98 & 38.37 & 35.84 & 39.45 & 30.57 & 32.21 & 32.55 \\
$M_t$ (single-stage) & 34.32 & 26.40 & 33.27 & 24.76 & 25.26 & 23.53 & 37.85 & 35.16 & 40.28 & 30.19 & 31.78 & 32.38 \\
$C_{23}\rightarrow C_t$ \cite{teng_2026pvsr} & 34.41 & 26.67 & 34.58 & 24.89 & 25.74 & 23.89 & 37.92 & 35.45 & 41.47 & 30.35 & 31.99 & 32.47 \\
$S_{23}\rightarrow S_t$ \cite{teng_2026pvsr} & 24.91 & 26.55 & 33.30 & 24.68 & 25.44 & 23.25 & 38.62 & 36.68 & 40.83 & 30.32 & 32.54 & 33.00 \\
$P_{23}\rightarrow P_t$ & 34.35 & 25.38 & 32.22 & 24.94 & 24.40 & 23.17 & 38.11 & 35.19 & 37.66 & 30.31 & 31.38 & 31.95 \\
$C_{23}\rightarrow P_t$ & 34.34 & 25.81 & 31.50 & 24.64 & 24.74 & 23.30 & 38.11 & 35.20 & 39.28 & 30.00 & 32.10 & 32.14 \\
$S_{23}\rightarrow P_t$ & 34.32 & 25.30 & 30.08 & 24.19 & 24.11 & 22.42 & 38.63 & 35.86 & 39.82 & 30.79 & 32.39 & 32.60 \\
$M_{23}\rightarrow M_t$ & 34.02 & 25.92 & 32.75 & 24.50 & 25.14 & 23.11 & 37.87 & 35.10 & 30.32 & 30.20 & 31.76 & 32.24 \\
$C_{23}\rightarrow S_{23}\rightarrow P_t$ & 34.43 & 25.68 & 31.14 & 24.53 & 24.49 & 23.02 & 38.11 & 35.20 & 39.18 & 29.88 & 31.96 & 32.07 \\
$\mathbf{S_{23}\rightarrow P_{23}\rightarrow P_t}$ & \textbf{33.44} & \textbf{24.35} & \textbf{29.07} & \textbf{23.63} & \textbf{23.95} & \textbf{22.21} & 37.84 & \textbf{34.25} & \textbf{36.08} & \textbf{29.71} & \textbf{31.18} & \textbf{31.28} \\
$P_{23}\rightarrow S_{23}\rightarrow C_t$ & 34.61 & 25.86 & 33.86 & 25.32 & 25.08 & 23.98 & 37.91 & 35.53 & 41.00 & 30.13 & 32.01 & 32.43 \\
\bottomrule
\end{tabular}
}
\end{table*}

Table~\ref{tab:curriculum} evaluates multiple curriculum training schedules for the proposed PECT framework. The K-fold pseudo-labels used during curriculum training are generated using the HP-VSR-ResFiLM model because it provides a good balance between phoneme recognition accuracy and realistic prediction errors. During inference, however, the same reconstruction model is evaluated using phoneme sequences generated by multiple Stage~1 VSR models, including V-ASR, PV-ASR, HP-VSR-Base, HP-VSR-ResFiLM, HP-VSR-FiLMFuse (L3--4), and HP-VSR-FiLMFuse (L4). This cross-model evaluation assesses whether the learned curriculum generalizes beyond the architecture used to generate pseudo-labels.

Among all curriculum schedules, the proposed $S_{23}\rightarrow P_{23}\rightarrow P_t$ consistently achieves the lowest WER across nearly all Stage~1 VSR models on both LRS2 and LRS3. Compared with single-stage training and alternative curriculum strategies, progressively adapting the reconstruction model from synthetic phoneme perturbations to multi-domain pseudo-labels, and finally to target-domain pseudo-labels, yields the most robust phoneme-to-text reconstruction. The consistent improvements observed for V-ASR, PV-ASR, and all HP-VSR variants indicate that the proposed curriculum learns error-correction patterns that generalize across different phoneme prediction models rather than overfitting to a specific VSR architecture.

The results further demonstrate that the order of the curriculum is critical. Although all three stages use noisy phoneme supervision, reversing the curriculum ($P_{23}\rightarrow S_{23}\rightarrow C_t$) substantially degrades performance, approaching the accuracy of single-stage training. This suggests that starting with synthetic phoneme perturbations yields a stable initialization before exposing the reconstruction model to increasingly realistic pseudo-label distributions. Premature adaptation to highly noisy pseudo-labels followed by cleaner supervision disrupts this progressive learning process and reduces the model's ability to generalize.

The mixed single-stage strategy ($M_t$) further highlights the importance of curriculum scheduling. Although $M_t$ exposes the reconstruction model to all available supervision types, including clean, synthetic perturbations and pseudo-labels from multiple domains, all samples are presented simultaneously without considering their relative difficulty or realism. Consequently, it consistently underperforms the proposed $S_{23}\rightarrow P_{23}\rightarrow P_t$ curriculum across all evaluated Stage~1 VSR models. This finding indicates that performance gains cannot be attributed solely to increased data diversity, but instead arise from progressively organizing supervision according to the realism of phoneme prediction errors.

In addition, we evaluate a two-stage mixed training strategy ($M_{23}\rightarrow M_t$), where clean phoneme sequences, synthetic perturbations, and pseudo-labels are randomly sampled with equal probability during the first stage, followed by target-domain adaptation. Compared with the single-stage mixed strategy, this schedule provides a modest improvement, suggesting that additional target-domain adaptation is beneficial. However, it still consistently underperforms the proposed progressive curriculum across most Stage~1 VSR models. These results indicate that simply exposing the reconstruction model to all supervision sources, even over multiple training stages, is insufficient. Instead, the gradual transition from easier synthetic perturbations to increasingly realistic pseudo-labels is critical for learning robust and generalizable error correction patterns.

Interestingly, the final target-domain adaptation provides different levels of improvement on the two benchmarks. On LRS3, target-domain pseudo-labels consistently yield the best performance, indicating that additional adaptation effectively captures dataset-specific phoneme error characteristics. On LRS2, the improvement over the multi-domain pseudo-label stage is smaller, suggesting that the combined LRS2 and LRS3 pseudo-labels already provide a close approximation to the target-domain error distribution. Collectively, these results demonstrate that the success of PECT stems from the progressive ordering of supervision rather than the individual supervision sources themselves.

\begin{table*}[!t]
\centering
\caption{Qualitative examples of phoneme-to-text reconstruction under different curriculum strategies. Differences from the ground-truth phoneme sequence and errors in the reconstructed sentences are highlighted. Red denotes incorrect phonemes or words, while blue denotes the corresponding ground-truth phonemes or words.}
\label{tab:qualitative_examples}
\small
\begin{tabular}{p{3.4cm} p{11.2cm}}
\toprule

\multicolumn{2}{c}{\textbf{Example 1: Phoneme Ambiguity}} \\
\midrule
\textbf{Predicted Phonemes} &
IH N \textcolor{red}{AH} F AO R M AH V AH D IH Z ER T \textcolor{red}{T UW} P L EY \\
\textbf{Ground Truth Phonemes} &
IH N {DH AH} F AO R M AH V AH D IH Z ER T P L EY {T} \\
\textbf{Target Sentence} &
In {the} form of a dessert {plate}. \\
\midrule
$C_{23}\rightarrow C_t$ &
In the form of a dessert \textcolor{red}{to play}. \\
$S_{23}\rightarrow S_t$ &
In \textcolor{red}{a} form of a dessert \textcolor{red}{to play}. \\
$\mathbf{S_{23}\rightarrow P_{23}\rightarrow P_t}$ &
In the form of a dessert plate. \\
\midrule

\multicolumn{2}{c}{\textbf{Example 2: Context-Aware Word Recovery}} \\
\midrule
\textbf{Predicted Phonemes} &
AE Z W IY AH P \textcolor{red}{AO R T T UW} R IH M EH M B R AH N S S AH N D EY \\
\textbf{Ground Truth Phonemes} &
AE Z W IY AH P {R OW CH} R IH M EH M B R AH N S S AH N D EY \\
\textbf{Target Sentence} &
As we {approach Remembrance} Sunday. \\
\midrule
$C_{23}\rightarrow C_t$ &
As we \textcolor{red}{support to remember one} Sunday. \\
$S_{23}\rightarrow S_t$ &
As we \textcolor{red}{apport} to Remembrance Sunday. \\
$\mathbf{S_{23}\rightarrow P_{23}\rightarrow P_t}$ &
As we approach Remembrance Sunday. \\
\midrule

\multicolumn{2}{c}{\textbf{Example 3: Missing Function Word}} \\
\midrule
\textbf{Predicted Phonemes} &
L EH T S T EY K AH L UH K B IH HH AY N D DH AH S IY N Z HH AW IH T AO L G EY M \\
\textbf{Ground Truth Phonemes} &
L EH T S T EY K AH L UH K B IH HH AY N D DH AH S IY N Z {AE T} HH AW IH T AO L {K} EY M \\
\textbf{Target Sentence} &
Let's take a look behind the scenes {at} how it all {came}. \\
\midrule
$C_{23}\rightarrow C_t$ &
Let's take a look behind the scenes how it all came. \\
$S_{23}\rightarrow S_t$ &
Let's take a look behind the scenes how it all \textcolor{red}{game}. \\
$\mathbf{S_{23}\rightarrow P_{23}\rightarrow P_t}$ &
Let's take a look behind the scenes at how it all came. \\
\midrule

\multicolumn{2}{c}{\textbf{Example 4: Phrase Completion}} \\
\midrule
\textbf{Predicted Phonemes} &
R AH G AA R D Z W AH T HH AE P AH N Z \\
\textbf{Ground Truth Phonemes} &
R AH G AA R D {L AH S AH V} W AH T HH AE P AH N Z \\
\textbf{Target Sentence} &
{Regardless of} what happens. \\
\midrule
$C_{23}\rightarrow C_t$ &
Regardless what happens. \\
$S_{23}\rightarrow S_t$ &
\textcolor{red}{Regards} what happens. \\
$\mathbf{S_{23}\rightarrow P_{23}\rightarrow P_t}$ &
Regardless of what happens. \\
\midrule

\multicolumn{2}{c}{\textbf{Example 5: Singular vs. Plural Recovery}} \\
\midrule
\textbf{Predicted Phonemes} &
DH EH R Z N OW AA B V IY AH S R IY \textcolor{red}{S AH N T S} \\
\textbf{Ground Truth Phonemes} &
DH EH R Z N OW AA B V IY AH S R IY {Z AH N} \\
\textbf{Target Sentence} &
There's no obvious {reason}. \\
\midrule
$C_{23}\rightarrow C_t$ &
There's no obvious \textcolor{red}{reasons}. \\
$S_{23}\rightarrow S_t$ &
There's no obvious \textcolor{red}{recents}. \\
$\mathbf{S_{23}\rightarrow P_{23}\rightarrow P_t}$ &
There's no obvious reason. \\

\midrule
\multicolumn{2}{c}{\textbf{Example 6: Hallucinated Word Insertion}} \\
\midrule
\textbf{Predicted Phonemes} &
S AH M T AY M Z IH T D AH Z \textcolor{red}{AA N} \\
\textbf{Ground Truth Phonemes} &
S AH M T AY M Z IH T D AH Z {AH N T} \\
\textbf{Target Sentence} &
Sometimes it {doesn't}. \\
\midrule
$C_{23}\rightarrow C_t$ &
Sometimes it \color{red}{does come on}. \\
$S_{23}\rightarrow S_t$ &
Sometimes it \color{red}{does on}. \\
$\mathbf{S_{23}\rightarrow P_{23}\rightarrow P_t}$ &
Sometimes it \color{red}{does}. \\
\bottomrule
\end{tabular}
\end{table*}

\subsection{Qualitative Analysis}

The quantitative results presented in Tables~\ref{tab:sota_curriculum}--\ref{tab:curriculum} report the average WER over all test utterances. To further understand the improvements introduced by the proposed PECT, Table~\ref{tab:qualitative_examples} presents representative examples of phoneme-to-text reconstruction under different curriculum strategies. The examples are categorized according to common reconstruction challenges, including phoneme ambiguity, context-aware word recovery, missing function words, phrase completion, singular--plural ambiguity, and hallucinated word insertion.

Examples~1--5 demonstrate that PECT-NLLB more effectively exploits linguistic context to recover the intended sentence despite imperfect phoneme inputs. In Example~1, the proposed method correctly reconstructs ``\emph{the}'' and ``\emph{plate}'' from ambiguous phoneme predictions, whereas the baseline curriculum incorrectly generates the phrase ``to play.'' Example~2 shows that PECT-NLLB resolves multiple corrupted phonemes to recover the semantically coherent phrase ``approach Remembrance''. At the same time, the baseline models produce unrelated or incomplete words such as ``support to remember one'' and ``apport.'' In Example~3, PECT-NLLB restores the missing function word ``at'' and correctly predicts ``came''. In contrast, baseline methods either omit the preposition or incorrectly reconstruct ``game.'' Example~4 illustrates successful phrase completion by recovering the complete expression ``regardless of,'' despite several missing phonemes. Similarly, in Example~5, PECT-NLLB correctly distinguishes the singular noun ``reason'' from the incorrect plural or phonetically similar alternatives generated by the baseline models.

Example~6 highlights a remaining limitation of the proposed approach. Although PECT-NLLB successfully suppresses the hallucinated word insertions (``come on'' and ``on'') produced by the baseline curricula, it fails to recover the missing negation, reconstructing ``does'' instead of ``doesn't.'' This example suggests that, while the proposed curriculum substantially improves robustness to phoneme recognition errors and reduces hallucinated word generation, certain semantic errors caused by severely corrupted or incomplete phoneme sequences remain difficult.

In summary, the qualitative analysis supports the quantitative findings by showing that the proposed curriculum improves the reconstruction model's ability to handle realistic phoneme prediction errors rather than simply memorizing clean phoneme-to-text mappings. The consistent recovery of correct words and sentence structures across diverse error patterns indicates that PECT-NLLB learns a more robust mapping from noisy phoneme sequences to natural language.

\section{Limitations and Future Work}
\label{sec:limitations}

Although the proposed PECT framework improves phoneme-to-text reconstruction under realistic phoneme prediction errors, several limitations remain, offering opportunities for future research.

\begin{itemize}

\item \textbf{Dependence on Stage~1 phoneme quality.}
The proposed framework operates only at the phoneme-to-text reconstruction stage, so it still depends on the quality of the phoneme sequences produced by the VSR frontend. Although PECT-NLLB improves robustness to moderate phoneme prediction errors, its ability to recover the target sentence is fundamentally limited by the information preserved in the predicted phoneme sequence. When critical phonetic information, such as negation or multiple consecutive phonemes, is missing or incorrectly recognized, the intended semantic content may no longer be recoverable, as shown in the qualitative examples. Future work will explore joint optimization of the VSR and reconstruction model, and will incorporate confidence estimation or n-best phoneme hypotheses to reduce error propagation and provide richer information for sentence reconstruction.

\item \textbf{Limited language coverage.}
All experiments are conducted on English-language datasets using ARPAbet phoneme representations. While NLLB is inherently multilingual, the effectiveness of the proposed curriculum learning strategy has not been evaluated on multilingual or cross-lingual VSR tasks. Future work will extend the framework to additional languages and phoneme inventories to assess its generalizability.

\item \textbf{Simplified synthetic error modeling.}
The synthetic error injector uses random insertion, deletion, and substitution operations to simulate phoneme prediction errors. Although this approach provides effective augmentation, it does not explicitly model phonetic similarity, viseme confusion, or context-dependent error patterns commonly observed in practical VSR systems. Future work may investigate linguistically informed or learned error generation methods capable of producing more realistic phoneme perturbations.

\end{itemize}

Taken together, PECT-NLLB demonstrates that progressively adapting phoneme-to-text reconstruction models from clean phoneme supervision to realistic VSR-generated errors is a promising strategy for reducing the training--inference mismatch in phoneme-centric VSR systems.

\section{Conclusion}
\label{sec:conlcusion}
This paper presents PECT, a curriculum learning framework that adapts the NLLB model for phoneme-to-text reconstruction in phoneme-centric visual speech recognition. By progressively training the reconstruction model with synthetic phoneme perturbations, source-domain pseudo-labels, and target-domain VSR-generated pseudo-labels, PECT effectively reduces the mismatch between training and inference, resulting in more robust phoneme-to-text reconstruction. Experimental results on the LRS2 and LRS3 benchmarks show that PECT consistently improves reconstruction performance across multiple Stage~1 VSR models and achieves the best performance when combined with the proposed HP-VSR variants. Qualitative analyses further show that PECT produces more accurate and linguistically coherent sentence reconstructions by correcting phoneme ambiguities, recovering missing words, and reducing hallucinated word insertions. These findings suggest that progressively adapting the reconstruction model to increasingly realistic phoneme prediction errors is an effective strategy for improving phoneme-centric visual speech recognition. Future work will investigate parameter-efficient adaptation techniques for large language models and extend the proposed curriculum framework to multilingual and audio-visual speech recognition tasks.

\bibliographystyle{IEEEtran}
\bibliography{sn_bibliography}

@String(CVPR= {IEEE Conf. Comput. Vis. Pattern Recog.})

@String(ICASSP=	{ICASSP})

@String(ICLR = {Int. Conf. Learn. Represent.})

@String(AAAI = {AAAI})

@String(CVPR  = {CVPR})

@String(ICLR  = {ICLR})

@inproceedings{xie_2020noisystudent,
  title     = {Self-training with noisy student improves imagenet classification},
  author    = {Xie, Qizhe and Luong, Minh-Thang and Hovy, Eduard and Le, Quoc V},
  booktitle = {Proceedings of the IEEE/CVF conference on computer vision and pattern recognition},
  pages     = {10687--10698},
  year      = {2020}
}

@article{sohn_2020fixmatch,
  title={Fixmatch: Simplifying semi-supervised learning with consistency and confidence},
  author={Sohn, Kihyuk and Berthelot, David and Carlini, Nicholas and Zhang, Zizhao and Zhang, Han and Raffel, Colin and Cubuk, Ekin Dogus and Kurakin, Alexey and Li, Chun-Liang},
  journal={Advances in neural information processing systems},
  volume={33},
  pages={596--608},
  year={2020}
}

@inproceedings{chen_2021cps,
  title={Semi-supervised semantic segmentation with cross pseudo supervision},
  author={Chen, Xiaokang and Yuan, Yuhui and Zeng, Gang and Wang, Jingdong},
  booktitle={Proceedings of the IEEE/CVF conference on computer vision and pattern recognition},
  pages={2613--2622},
  year={2021}
}

@inproceedings{xu_2021softteacher,
  title={End-to-end semi-supervised object detection with soft teacher},
  author={Xu, Mengde and Zhang, Zheng and Hu, Han and Wang, Jianfeng and Wang, Lijuan and Wei, Fangyun and Bai, Xiang and Liu, Zicheng},
  booktitle={Proceedings of the IEEE/CVF international conference on computer vision},
  pages={3060--3069},
  year={2021}
}

@article{shi2024robustger,
  title       = {Large Language Models are Efficient Learners of Noise-Robust Speech Recognition},
  author      = {Hu, Yuchen and Chen, Chen and Yang, Chao-Han Huck and Li, Ruizhe and Zhang, Chao and Chen, Pin-Yu and Chng, Eng Siong},
  journal     = {Proceedings of the International Conference on Learning Representations (ICLR)},
  year        = {2024}
}

@article{hari2024curriculum,
  title       = {Curriculum Learning for Cross-Lingual Data-to-Text Generation With Noisy Data},
  author      = {Hari, Kancharla Aditya and Gupta, Manish and Varma, Vasudeva},
  journal     = {arXiv preprint arXiv:2409.06714},
  year        = {2024}
}

@inproceedings{kumar2010selfpaced,
  title     = {Self-Paced Learning for Latent Variable Models},
  author    = {Kumar, M Pawan and Packer, Benjamin and Koller, Daphne},
  booktitle = {Advances in Neural Information Processing Systems (NeurIPS)},
  volume    = {23},
  year      = {2010}
}

@inproceedings{jiang2015selfpaced,
  title     = {Self-Paced Curriculum Learning},
  author    = {Jiang, Lu and Meng, Deyu and Mitamura, Teruko and Hauptmann, Alexander G},
  booktitle = {Proceedings of the AAAI Conference on Artificial Intelligence},
  volume    = {29},
  number    = {1},
  year      = {2015}
}

@inproceedings{graves2017automated,
  title     = {Automated Curriculum Learning for Neural Networks},
  author    = {Graves, Alex and Bellemare, Marc G and Menick, Jacob and Munos, R{\'e}mi and Kavukcuoglu, Koray},
  booktitle = {Proceedings of the 34th International Conference on Machine Learning (ICML)},
  pages     = {1311--1320},
  year      = {2017}
}

@inproceedings{hacohen2019power,
  title     = {On the Power of Curriculum Learning in Training Deep Networks},
  author    = {Hacohen, Guy and Weinshall, Daphna},
  booktitle = {Proceedings of the 36th International Conference on Machine Learning (ICML)},
  pages     = {2535--2544},
  year      = {2019}
}

@article{soviany2022curriculum,
  title     = {Curriculum Learning: A Survey},
  author    = {Soviany, Petru-Daniel and Ionescu, Radu Tudor and 
          Rota, Paolo and Sebe, Nicu},
  journal   = {International Journal of Computer Vision},
  volume    = {130},
  number    = {6},
  pages     = {1526--1565},
  year      = {2022},
  publisher = {Springer}
}

@inproceedings{peymanfard_2022ieee,
  title     = {Lip reading using external viseme decoding},
  author    = {Peymanfard, Javad and Mohammadi, Mohammad Reza and Zeinali, Hossein and Mozayani, Nasser},
  booktitle = {2022 International Conference on Machine Vision and Image Processing (MVIP)},
  pages     = {1--5},
  year      = {2022},
  organization  = {IEEE}
}

@inproceedings{binici2025medsage,
  title     = {MEDSAGE: enhancing robustness of medical dialogue summarization to ASR errors with llm-generated synthetic dialogues},
  author    = {Binici, Kuluhan and Kashyap, Abhinav Ramesh and Schlegel, Viktor and Liu, Andy T and Dwivedi, Vijay Prakash and Nguyen, Thanh-Tung and Gao, Xiaoxue and Chen, Nancy F and Winkler, Stefan},
  booktitle = {Proceedings of the AAAI Conference on Artificial Intelligence},
  volume    = {39},
  number    = {22},
  pages     = {23496--23504},
  year      = {2025}
}

@article{nakagome2022interaug,
  title     = {InterAug: augmenting noisy intermediate predictions for CTC-based ASR},
  author    = {Nakagome, Yu and Komatsu, Tatsuya and Fujita, Yusuke and Ichimura, Shuta and Kida, Yusuke},
  journal   = {arXiv preprint arXiv:2204.00174},
  year      = {2022}
}

@inproceedings{shillingford_2019interspeech,
  title     = {Large-Scale Visual Speech Recognition},
  author    = {Shillingford, Brendan and Assael, Yannis and Hoffman, Matthew W and Paine, Thomas and Hughes, C{\'\i}an and Prabhu, Utsav and Liao, Hank and Sak, Hasim and Rao, Kanishka and Bennett, Lorrayne and others},
  booktitle = {Proc. Interspeech 2019},
  pages     = {4135--4139},
  year      = {2019}
}

@inproceedings{chan_2016icassp,
  title     = {Listen, attend and spell: A neural network for large vocabulary conversational speech recognition},
  author    = {Chan, William and Jaitly, Navdeep and Le, Quoc and Vinyals, Oriol},
  booktitle = {2016 IEEE international conference on acoustics, speech and signal processing (ICASSP)},
  pages     = {4960--4964},
  year      = {2016},
  organization  = {IEEE}
}

@article{watanabe_2018espnet,
  title     = {ESPnet: End-to-End Speech Processing Toolkit},
  author    = {Watanabe, Shinji and Hori, Takaaki and Karita, Shigeki and Hayashi, Tomoki and Nishitoba, Jiro and Unno, Yuya and Soplin, Nelson Enrique Yalta and Heymann, Jahn and Wiesner, Matthew and Chen, Nanxin and others},
  journal   = {Interspeech 2018},
  year      = {2018},
  publisher = {ISCA}
}

@inproceedings{lee_2023asru,
  title     = {Optimizing two-pass cross-lingual transfer learning: Phoneme recognition and phoneme to grapheme translation},
  author    = {Lee, Wonjun and Lee, Gary Geunbae and Kim, Yunsu},
  booktitle = {2023 IEEE Automatic Speech Recognition and Understanding Workshop (ASRU)},
  pages     = {1--8},
  year      = {2023},
  organization  = {IEEE}
}

@inproceedings{zhang_2023speechgpt,
  title     = {Speechgpt: Empowering large language models with intrinsic cross-modal conversational abilities},
  author    = {Zhang, Dong and Li, Shimin and Zhang, Xin and Zhan, Jun and Wang, Pengyu and Zhou, Yaqian and Qiu, Xipeng},
  booktitle = {Findings of the Association for Computational Linguistics: EMNLP 2023},
  pages     = {15757--15773},
  year      = {2023}
}

@inproceedings{radford_2023whisper,
  title     = {Robust speech recognition via large-scale weak supervision},
  author    = {Radford, Alec and Kim, Jong Wook and Xu, Tao and Brockman, Greg and McLeavey, Christine and Sutskever, Ilya},
  booktitle = {International conference on machine learning},
  pages     = {28492--28518},
  year      = {2023},
  organization = {PMLR}
}

@article{bear_2017SC,
  title     = {Phoneme-to-viseme mappings: the good, the bad, and the ugly},
  author    = {Bear, Helen L and Harvey, Richard},
  journal   = {Speech Communication},
  volume    = {95},
  pages     = {40--67},
  year      = {2017},
  publisher = {Elsevier}
}

@article{cui_2021asr,
  title     = {An approach to improve robustness of nlp systems against asr errors},
  author    = {Cui, Tong and Xiao, Jinghui and Li, Liangyou and Jiang, Xin and Liu, Qun},
  journal   = {arXiv preprint arXiv:2103.13610},
  year      = {2021}
}

@inproceedings{wang_2020nlpcai,
  title     = {Data augmentation for training dialog models robust to speech recognition errors},
  author    = {Wang, Longshaokan and Fazel-Zarandi, Maryam and Tiwari, Aditya and Matsoukas, Spyros and Polymenakos, Lazaros},
  booktitle = {Proceedings of the 2nd Workshop on Natural Language Processing for Conversational AI},
  pages     = {63--70},
  year      = {2020}
}

@article{jin_2022filterevolve,
  title     = {Filter and evolve: progressive pseudo label refining for semi-supervised automatic speech recognition},
  author    = {Jin, Zezhong and Zhong, Dading and Song, Xiao and Liu, Zhaoyi and Ye, Naipeng and Zeng, Qingcheng},
  journal   = {arXiv preprint arXiv:2210.16318},
  year      = {2022}
}

@inproceedings{shi_2022avhubert,
  author    = {Bowen Shi and Wei-Ning Hsu and Kushal Lakhotia 
               and Abdelrahman Mohamed},
  title     = {Learning Audio-Visual Speech Representation by Masked 
               Multimodal Cluster Prediction},
  booktitle = {International Conference on Learning Representations 
               ({ICLR})},
  year      = {2022}
}

@inproceedings{shi_2022interspeech,
  author    = {Bowen Shi and Wei-Ning Hsu and Abdelrahman Mohamed},
  title     = {Robust Self-Supervised Audio-Visual Speech Recognition},
  booktitle = {Interspeech},
  year      = {2022}
}

@inproceedings{prajwal_2024interspeech,
  author    = {K R Prajwal and Triantafyllos Afouras 
               and Andrew Zisserman},
  title     = {Speech Recognition Models are Strong Lip-Readers},
  booktitle = {Interspeech},
  pages     = {2425--2429},
  doi       = {10.21437/Interspeech.2024-2290},
  year      = {2024}
}

@inproceedings{cappellazzo_2025llama,
  title     = {Large language models are strong audio-visual speech recognition learners},
  author    = {Cappellazzo, Umberto and Kim, Minsu and Chen, Honglie and Ma, Pingchuan and Petridis, Stavros and Falavigna, Daniele and Brutti, Alessio and Pantic, Maja},
  booktitle = {ICASSP 2025-2025 IEEE International Conference on Acoustics, Speech and Signal Processing (ICASSP)},
  pages     = {1--5},
  year      = {2025},
  organization  = {IEEE}
}

@inproceedings{bengio_2009curriculum,
  title     = {Curriculum learning},
  author    = {Bengio, Yoshua and Louradour, J{\'e}r{\^o}me and Collobert, Ronan and Weston, Jason},
  booktitle = {Proceedings of the 26th annual international conference on machine learning},
  pages     = {41--48},
  year      = {2009}
}

@article{bengio_2015scheduled,
  title     = {Scheduled sampling for sequence prediction with recurrent neural networks},
  author    = {Bengio, Samy and Vinyals, Oriol and Jaitly, Navdeep and Shazeer, Noam},
  journal   = {Advances in neural information processing systems},
  volume    = {28},
  year      = {2015}
}

@article{song_2023ieee,
  title     = {Learning From Noisy Labels With Deep Neural Networks: A Survey},
  author    = {Song, Hwanjun and Kim, Minseok and Park, Dongmin and Shin, Yooju and Lee, Jae-Gil},
  journal   = {IEEE transactions on neural networks and learning systems},
  volume    = {34},
  number    = {11},
  pages     = {8135--8153},
  year      = {2023}
}

@article{teng_2026hpvsr,
title   = {Head-Pose-Aware Visual Speech Recognition via Residual FiLM Modulation},
author  = {Teng, Matthew Kit Khinn and Zhang, Haibo and Saitoh, Takeshi},
journal = {arXiv preprint arXiv:2606.00751},
year    = {2026}
}

@inproceedings{teng_2026pvsr,
  title     = {Phoneme-Level Visual Speech Recognition via Point-Visual Fusion and Language Model Reconstruction},
  author    = {Teng, Matthew Kit Khinn and Zhang, Haibo and Saitoh, Takeshi},
  booktitle = {ICASSP 2026-2026 IEEE International Conference on Acoustics, Speech and Signal Processing (ICASSP)},
  pages     = {10477--10481},
  year      = {2026},
  organization  = {IEEE}
}

@inproceedings{son_2017lrs2,
    title   = "Lip reading sentences in the wild",
    author  = "Joon Son Chung and Andrew Senior and Oriol Vinyals and Andrew Zisserman",
    booktitle = "IEEE Conference on Computer Vision and Pattern Recognition (CVPR)",
    pages   = "3444--3453",
    year    = "2017",
    doi     = "10.1109/CVPR.2017.367"
}

@article{afouras_2018lrs3,
    title   = "{LRS3-TED}: a large-scale dataset for visual speech recognition",
    author  = "Triantafyllos Afouras and Joon Son Chung and Andrew Zisserman",
    journal = "arXiv preprint arXiv:1809.00496",
    year    = "2018",
    doi     = "10.48550/arXiv.1809.00496"
}

@inproceedings{thomas_2025vallr,
  title     = "Vallr: Visual asr language model for lip reading",
  author    = "Thomas, Marshall and Fish, Edward and Bowden, Richard",
  booktitle = "Proceedings of the IEEE/CVF International Conference on Computer Vision",
  pages     = "2846--2856",
  year      = "2025"
}

@inproceedings{ma_2023avsr,
    title   = "{Auto-AVSR}: Audio-Visual Speech Recognition with Automatic Labels",
    author  = "Pingchuan Ma and Alexandros Haliassos and Adriana Fernandez-Lopez and Honglie Chen and Stavros Petridis and Maja Pantic",
    booktitle = "IEEE International Conference on Acoustics, Speech and Signal Processing (ICASSP)",
    pages   = "1--5",
    year    = "2023",
    doi     = "10.1109/ICASSP49357.2023.10096889"
}

@inproceedings{prajwal_2022subword,
    title   = "Sub-word Level Lip Reading With Visual Attention",
    author  = "K R Prajwal and Triantafyllos Afouras and Andrew Zisserman",
    booktitle = "IEEE/CVF Conference on Computer Vision and Pattern Recognition (CVPR)",
    pages   = "5162--5172",
    year    = "2022",
    doi     = "10.1109/CVPR52688.2022.00510"
}

@article{elBialy_CAAI_2023,
    title   = "{Developing phoneme-based lip-reading sentences system for silent speech recognition}",
    author  = "Randa El‐Bialy and Daqing Chen and Souheil Fenghour and Walid Hussein and Perry Xiao and Omar H. Karam and Bo Li",
    journal = "CAAI Transactions on Intelligence Technology",
    volume  = "8",
    number  = "1",
    pages   = "129--138",
    year    = "2023",
    doi     = "10.1049/cit2.12131"
}

@inproceedings{zhang_2024nemo,
    title   = "A Chat about Boring Problems: Studying {GPT}-Based Text Normalization",
    author  = "Yang Zhang and Travis M. Bartley and Mariana Graterol-Fuenmayor and Vitaly Lavrukhin and Evelina Bakhturina and Boris Ginsburg",
    booktitle = "IEEE International Conference on Acoustics, Speech and Signal Processing (ICASSP)",
    pages   = "10921--10925",
    year    = "2024",
    doi     = "10.1109/ICASSP48485.2024.10447169"
}

@inproceedings{ploujnikov_2022soundchoice,
    title   = "{SoundChoice}: Grapheme-to-Phoneme Models with Semantic Disambiguation",
    author  = "Artem Ploujnikov and Mirco Ravanelli",
    booktitle = "INTERSPEECH",
    pages   = "486--490",
    year    = "2022",
    doi     = "10.21437/Interspeech.2022-11066"
}

@article{costa_2022nllb,
    title   = "No Language Left Behind: Scaling Human-Centered Machine Translation",
    author  = "Marta R. Costa-jussà and James Cross and Onur Çelebi and Maha Elbayad and Kenneth Heafield and Kevin Heffernan and Elahe Kalbassi and Janice Lam and Daniel Licht and Jean Maillard and Anna Sun and Skyler Wang and Guillaume Wenzek and Al Youngblood and Bapi Akula and Loic Barrault and Gabriel Mejia Gonzalez and Prangthip Hansanti and John Hoffman and Semarley Jarrett and Kaushik Ram Sadagopan and Dirk Rowe and Shannon Spruit and Chau Tran and Pierre Andrews and Necip Fazil Ayan and Shruti Bhosale and Sergey Edunov and Angela Fan and Cynthia Gao and Vedanuj Goswami and Francisco Guzmán and Philipp Koehn and Alexandre Mourachko and Christophe Ropers and Safiyyah Saleem and Holger Schwenk and Jeff Wang ",
    journal = "arXiv preprint arXiv:2207.04672",
    year    = "2022",
    doi     = "10.48550/arXiv.2207.04672"
}

@article{jelinek_1975wer,
    title   = "Design of a linguistic statistical decoder for the recognition of continuous speech",
    author  = "Jelinek, Frederick and Bahl, Lalit and Mercer, Robert",
    journal = "IEEE Transactions on Information Theory",
    volume  = "21",
    number  = "3",
    pages   = "250--256",
    year    = "1975",
    doi     = "10.1109/TIT.1975.1055384"
}

\EOD

\end{document}